\documentclass[journal]{IEEEtran}
\usepackage{cite}
\usepackage{amsmath,amssymb,amsfonts}
\usepackage{algorithmic}
\usepackage{booktabs}
\usepackage{makecell}
\usepackage{multirow}

\usepackage{enumitem}
\usepackage{tabularx} 
\usepackage{graphicx}
\usepackage{algorithm,algorithmic}
\usepackage{float} 
\usepackage{hyperref}
\hypersetup{hidelinks}
\usepackage{textcomp}
\newcounter{subfigctr}[figure]
\renewcommand{\thesubfigctr}{\alph{subfigctr}}

\newcommand{\mysubfig}[3]{%
    \begin{minipage}[t]{0.32\textwidth}
        \centering
        \refstepcounter{subfigctr}
        \includegraphics[width=\linewidth]{#1}

        \vspace{2pt}
        {\footnotesize (\thesubfigctr)\ #2\label{#3}}
    \end{minipage}
}
\newcommand{\secbest}[1]{\underline{\rule{0pt}{1.4ex}#1}}
\newcommand{\mstd}[2]{\mbox{#1$\pm$#2}} 

\hypersetup{
    colorlinks=true,
    linkcolor=blue,
    filecolor=blue,
    urlcolor=blue,
    citecolor=blue
}
\def\BibTeX{{\rm B\kern-.05em{\sc i\kern-.025em b}\kern-.08em
    T\kern-.1667em\lower.7ex\hbox{E}\kern-.125emX}}
\begin{document}
\title{Distance-Aware Joint Spatio-Temporal Graph Contrastive Learning for Major Depressive Disorder Diagnosis}
\author{Muhammad Asif Hasan, Yanming Zhu, Xuefei Yin, and Alan Wee-Chung Liew, \IEEEmembership{Senior Member, IEEE}
\thanks{Corresponding author: Alan Wee-Chung Liew}
\thanks{Muhammad Asif Hasan, Yanming Zhu, Xuefei Yin, and Alan Wee-Chung Liew are with the School of Information Technology and Communication, Griffith University, Australia. (Email: a.liew@griffith.edu.au)}}

\maketitle

\begin{abstract}
Major depressive disorder (MDD) is a prevalent neuropsychiatric disorder, and accurate diagnosis from resting-state functional magnetic resonance imaging remains challenging. Dynamic functional connectivity (DFC) captures time-varying interactions among brain regions and provides rich spatio-temporal information, but existing DFC-based methods remain limited by three factors. First, sliding-window Pearson correlation often produces noisy and unstable connectivity estimates that are sensitive to window configuration and residual motion artifacts. Second, correlation-based node representations underexploit informative frequency-domain characteristics of regional blood-oxygen-level-dependent signals. Third, most spatio-temporal graph models treat spatial topology and temporal evolution separately, limiting their ability to capture the coupled dynamics of functional brain networks. 
To address these limitations, we reformulate DFC learning as joint spatio-temporal graph representation learning under an explicit Hawkes-inspired temporal dependency prior. 
Accordingly, we propose HWSTCL, a unified framework that learns DFC representations on a reliability-refined joint spatio-temporal graph with an aligned kernel-weighted pretraining objective. Specifically, regional blood-oxygen-level-dependent signals within each temporal window are encoded as spectral node descriptors, while functional edges are refined by an exponential distance-decay prior to suppress unreliable long-range associations. We then construct a joint spatio-temporal graph by connecting each region across future windows with a Hawkes-inspired exponential kernel, enabling spatial topology and temporal evolution to be learned within the same message-passing operator. On top of this joint graph, we introduce a Hawkes-inspired kernel-weighted contrastive objective that enforces region-anchored temporal consistency by emphasizing nearby cross-window pairs while suppressing redundant cross-region similarity. Experiments demonstrate that HWSTCL outperforms recent baselines and learns coherent spatio-temporal representations for MDD diagnosis.
\end{abstract}


\section{Introduction}
\label{sec:introduction}
\IEEEPARstart{M}{ajor} depressive disorder (MDD) is a prevalent neuropsychiatric disorder and a major contributor to disability worldwide \cite{marx2023major}. Despite its high clinical burden, MDD diagnosis still relies heavily on symptom-based assessment, which remains vulnerable to heterogeneity in clinical presentation and overlap with related psychiatric conditions. Resting-state functional magnetic resonance imaging (rs-fMRI) provides a non-invasive means of characterizing large-scale brain activity and therefore offers a promising basis for objective imaging-assisted diagnosis. However, learning reliable disease-related representations from rs-fMRI remains challenging, particularly when diagnostic signals are distributed across interacting brain regions.

A growing body of evidence suggests that MDD is associated with abnormal functional connectivity (FC) patterns \cite{zhang2021identification}. FC derived from blood-oxygen-level-dependent (BOLD) signals characterizes statistical dependencies among brain regions and has therefore become a useful substrate for biomarker discovery and disease classification. Recent deep learning studies have increasingly modeled FC as a graph, with regions of interest (ROIs) as nodes and inter-regional connectivity as edges, enabling graph neural networks (GNNs) to exploit the topology of brain networks for MDD diagnosis \cite{Noman2024,si2025graph}. Although these graph-based approaches improve over conventional vectorized connectivity analysis, many rely on Static FC (SFC) and therefore cannot characterize temporal variation in inter-regional interactions.

FC is now widely recognized as time-varying rather than stationary across an entire scan \cite{hindriks2016can}. Dynamic FC (DFC) provides a richer description of brain organization by capturing how connectivity patterns evolve over time, and abnormal dynamic properties have been reported in several brain disorders, including MDD \cite{zhao2024enhancing}. This has motivated recent spatio-temporal graph models that combine graph learning with temporal modeling for brain disorder diagnosis \cite{kong2021spatio,kim2021learning,wang2022spatio}. In parallel, self-supervised and contrastive learning have emerged as promising tools for exploiting the intrinsic structure of neuroimaging data beyond class labels alone. Nevertheless, current DFC-based methods still lack a reliable formulation for learning spatio-temporal graph representations under an explicit temporal dependency prior.

This gap is reflected in three related limitations. 
First, DFC is commonly constructed by applying sliding-window Pearson correlation coefficient (PCC) to windowed BOLD signals \cite{zhu2024temporal,li2025long}. Although widely used, PCC captures marginal rather than direct associations and becomes unstable in the short-window regime, where connectivity estimates are more sensitive to noise, residual motion artifacts, and preprocessing choices. PCC-derived connectomes may also exhibit distance-dependent bias, which can complicate interpretation and downstream learning \cite{korponay2024brain}. 
Second, graph-based diagnostic models often use correlation-derived descriptors at the node level, potentially underutilizing informative frequency-domain characteristics of regional BOLD signals \cite{duan2024spectral}.
Third, many existing spatio-temporal graph methods still model spatial topology and temporal evolution in separate stages, which limits their ability to characterize the coupled dynamics of brain networks. These issues suggest that reliable DFC learning requires more than improved graph architecture alone. It requires a representation learning framework that jointly addresses edge reliability, node informativeness, and temporally coupled message passing.

To address these limitations, we formulate DFC analysis as representation learning on a reliability-refined joint spatio-temporal graph governed by a Hawkes-inspired temporal dependency prior. Based on this formulation, we propose \textbf{HWSTCL}, a framework for learning spatio-temporal representations from windowed rs-fMRI graphs for MDD diagnosis. Within each temporal window, ROI signals are encoded as spectral node descriptors to capture frequency-specific regional dynamics. PCC-derived edges are refined with a distance-decay prior to suppress unreliable long-range associations that are more susceptible to short-window estimation noise. Across windows, the same ROI is connected through a Hawkes-inspired exponential kernel, enabling spatial topology and temporal evolution to be learned within the same spatio-temporal graph rather than through separate modules. On top of this graph, we introduce a Hawkes-inspired kernel-weighted contrastive objective that promotes ROI-anchored temporal consistency while suppressing redundant cross-ROI similarity. The model is first pretrained using this contrastive objective and is then further refined for downstream MDD diagnosis through joint optimization of the supervised classification objective and the contrastive objective.

The main contributions of this work are as follows:
\begin{enumerate}[label={\arabic*)}]
    \item We reformulate DFC-based MDD diagnosis as joint spatio-temporal graph representation learning on a reliability-refined brain graph under an explicit Hawkes-inspired temporal dependency prior, enabling spatial topology and temporal evolution to be modeled within a unified graph propagation framework.
    \item We propose HWSTCL, which constructs a reliability-refined joint spatio-temporal graph from spectral ROI descriptors and distance-aware functional connectivity, and further introduces a Hawkes-inspired kernel-weighted contrastive objective that is aligned with the same temporal dependency prior for representation learning.
    \item We conduct extensive experiments on a benchmark rs-fMRI MDD dataset, and demonstrate that HWSTCL improves diagnostic performance while learning coherent spatio-temporal connectivity representations.
\end{enumerate}  

\section{Related Work}
\label{sec:R_W}

\subsection{Graph Learning for SFC and DFC}
Graph neural networks have been widely adopted for functional connectome analysis in neuropsychiatric disorders. Early graph-based studies primarily focused on SFC, where a single time-averaged connectome is used to characterize each subject. Representative examples include multi-atlas fusion for MDD diagnosis \cite{Lee2024}, graph autoencoder-based representation learning \cite{Noman2024}, and modular-attention-based graph modeling \cite{si2025graph}. These methods demonstrate the utility of graph representations for exploiting the topology of functional brain networks, but they do not account for temporal variation in inter-regional interactions.
To capture the non-stationary nature of functional connectivity, more recent studies have turned to DFC and spatio-temporal graph learning. Kong et al.~\cite{kong2021spatio} proposed a spatio-temporal graph convolutional network for MDD diagnosis, while Kim et al.~\cite{kim2021learning} and Wang et al.~\cite{wang2022spatio} combined graph learning with attention mechanisms for dynamic brain network modeling. Transformer-based architectures have also been introduced to model temporal dependencies across windows or segments \cite{wang2024leveraging,chen2023fe}. More recently, methods such as ESTA \cite{lee2024eigendecomposition}, Temporal-BCGCN \cite{zhu2024temporal}, MSSTAN \cite{kong2024multi}, and Long-Interval STGC \cite{li2025long} have further advanced dynamic graph modeling for rs-fMRI analysis. Despite these advances, many existing approaches still rely on separate spatial encoding and temporal aggregation stages, or they model temporal evolution without imposing an explicit dependency prior within graph construction itself. As a result, the coupled nature of spatial topology and temporal evolution in DFC may not be fully exploited, which can also hinder the stability of graph-based representation learning. 

\subsection{Reliability of DFC and Node Representation}
Reliable DFC construction remains a fundamental challenge in rs-fMRI analysis. For the diagnosis of neuropsychiatric disorders, FC (edges) is often considered more informative than brain regions (nodes) \cite{zheng2024brainib}. However, in most pipelines, DFC is estimated by partitioning BOLD time series into windows and computing inter-regional connectivity within each window, typically using the PCC \cite{zhu2024temporal,li2025long}. Although PCC is widely used because of its simplicity and interpretability, it captures marginal rather than direct associations and can become unstable when estimated from short windows. Such estimates are more sensitive to nuisance signals, preprocessing choices, and residual head motion \cite{smith2011network}. In particular, motion-related effects can induce distance-dependent bias in functional connectomes even after standard preprocessing, thereby complicating interpretation and downstream learning \cite{power2012spurious,korponay2024brain}.
Another underexplored issue is the choice of node representation. Graph-based diagnostic studies often emphasize connectivity patterns at the edge level and use correlation-derived or region-wise summary descriptors at the node level \cite{zheng2024brainib}. While connectivity alterations are central to psychiatric diagnosis and biomarker discovery \cite{insel2015brain,lozano2017waving,gallo2023functional,li2020neuroimaging}, correlation-based node descriptors may underutilize informative frequency-domain characteristics embedded in regional BOLD signals. Recent work suggests that spectral descriptors can provide complementary views of regional dynamics beyond correlation structure alone \cite{duan2024spectral}. However, their role in DFC-based graph learning for psychiatric diagnosis remains insufficiently investigated, particularly in conjunction with temporally evolving graph representations.

\subsection{Self-Supervised and Contrastive Learning for Functional Brain Networks}
Self-supervised learning has emerged as a promising strategy for exploiting the intrinsic structure of neuroimaging data beyond class labels alone. Among these approaches, contrastive learning has gained increasing attention in medical imaging and brain network analysis \cite{wang2025self}. For fMRI-based disorder classification, Zhang et al.~\cite{zhang2023gcl} applied graph contrastive learning to neurodevelopmental disorder diagnosis, Zhu et al.~\cite{zhu2022contrastive} proposed a contrastive multi-view framework for autism spectrum disorder classification, and Zhou et al.~\cite{zhou2025dclnet} introduced a collaborative contrastive learning strategy for brain disease analysis. These studies demonstrate that self-supervised objectives can improve the quality of learned brain representations.
Nevertheless, most existing self-supervised and contrastive methods for functional brain networks are designed for SFC graphs, graph-level consistency, or multi-view agreement. They do not explicitly target ROI-anchored temporal consistency in dynamic graphs, nor do they align temporal weighting in the contrastive objective with the dependency structure used for spatio-temporal graph construction. Consequently, the relationship between cross-window temporal coupling and self-supervised regularization remains weakly specified in prior work.

In summary, prior studies have advanced graph-based modeling and self-supervised learning for functional brain network analysis, but three aspects remain insufficiently addressed for DFC-based MDD diagnosis: reliable short-window graph construction, frequency-informed node representation, and a unified spatio-temporal learning framework in which temporal coupling and the auxiliary contrastive objective are governed by the same explicit dependency prior.

\begin{figure*}[htbp] 
\centerline{\includegraphics[width=1.0\textwidth]{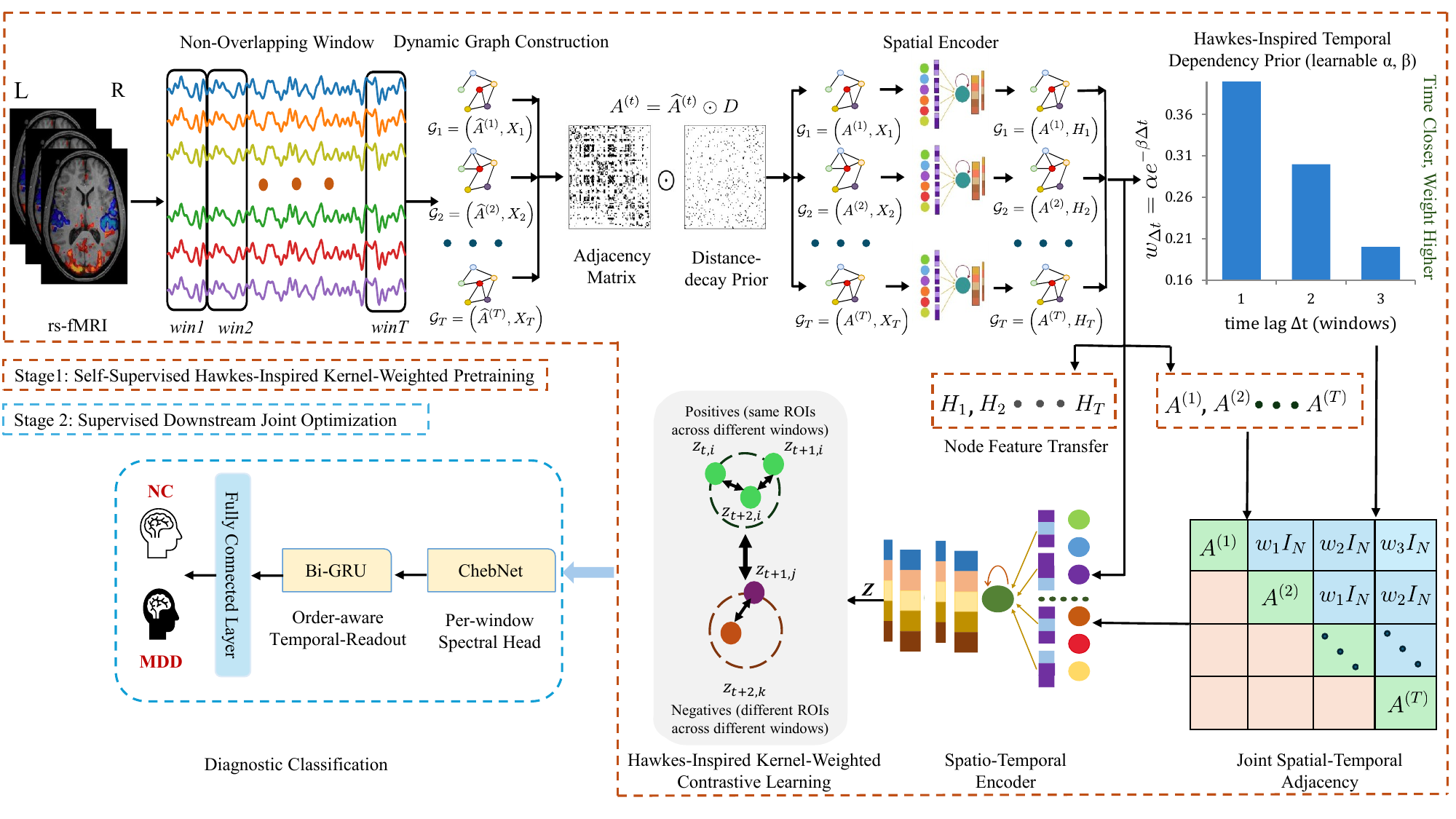}}
\caption{Overview of the proposed two-stage HWSTCL framework.
\textbf{Stage 1: Self-supervised Joint Spatio-Temporal Graph Pretraining.} The BOLD time series is partitioned into non-overlapping windows, and one functional graph is constructed for each window using FFT-derived ROI features and distance-refined functional connectivity. A local spatial encoder first lifts each graph to latent ROI representations, which are then organized into a joint spatio-temporal graph by introducing identity-preserving temporal links across future windows under a Hawkes-inspired exponential decay prior. The resulting weighted spatio-temporal encoder is pretrained using the HW-CL objective, which promotes temporally coherent and discriminative ROI representations by weighting cross-window positive and negative pairs according to temporal lag.
\textbf{Stage 2: Supervised Downstream Joint Optimization.} A diagnostic readout head composed of a ChebNet-based graph refinement layer, a bidirectional GRU, and a fully connected prediction layer is attached for subject-level classification. The pretrained encoder and downstream classifier are then jointly optimized under the supervised classification and HW-CL objectives.}
\label{fig1}
\end{figure*} 

\section{Proposed Method}
\label{sec:Proposed_Method}
As illustrated in Fig.~\ref{fig1}, we formulate DFC analysis as representation learning on a reliability-refined joint spatio-temporal graph governed by a Hawkes-inspired temporal dependency prior. Accordingly, we propose HWSTCL, a two-stage framework for MDD diagnosis from rs-fMRI BOLD signals. In Stage~1, windowed functional graphs are constructed from frequency-informed ROI descriptors and distance-refined connectivity estimates, and a weighted joint spatio-temporal graph encoder is pretrained using the HW-CL objective to learn discriminative ROI-level spatio-temporal representations. In Stage~2, a diagnostic readout head is attached, and the pretrained encoder is further refined together with the downstream classifier through joint optimization of the supervised classification objective and the HW-CL objective.

Importantly, the proposed scheme distinguishes between two roles. A window-wise spatial encoder first lifts each graph snapshot into a stable ROI-level latent space. The principal representation learning step is then carried out on a joint spatio-temporal graph, where within-window spatial structure and cross-window temporal couplings are propagated through the same weighted STGIN operator. In this way, the local spatial encoder serves as feature lifting, whereas the joint spatio-temporal graph and the HW-CL objective together define the main spatio-temporal representation learning mechanism.

\vspace{-0.1in}
\subsection{DFC Network Construction}
For each subject, the preprocessed rs-fMRI BOLD signal is parcellated into \(N=116\) ROIs using the Automated Anatomical Labelling (AAL) atlas \cite{Rolls2020}. The resulting ROI time series are divided into \(T\) non-overlapping windows of length \(S\) time points. Each window is converted into a graph snapshot \(\mathcal{G}_t=(V,E_t)\), where nodes correspond to ROIs and edges encode window-specific functional connectivity. To improve the reliability of short-window DFC estimation, we define both node features and edge weights in a manner tailored to the short-window regime.

\subsubsection{Node Features}
\label{sec:node_f}
Within each window \(t\in\{1,\dots,T\}\), let \(x_i^{(t)}\in\mathbb{R}^{S}\) denote the BOLD signal segment of the \(i\)-th ROI in window \(t\). For every ROI, we compute the one-sided fast Fourier transform (FFT) of \(x_i^{(t)}\) and retain coefficients in the low-frequency band \(\mathcal{B}=[0.01,0.10]\) Hz, which is the frequency range most commonly associated with resting-state BOLD fluctuations after standard preprocessing \cite{yan2010dparsf}. We use the log-magnitude spectrum rather than phase because phase estimates from short BOLD segments are highly sensitive to window boundaries and are therefore less stable in the short-window setting used for DFC. In contrast, the magnitude spectrum provides a compact summary of regional oscillatory content while remaining computationally simple and parameter-free. The sensitivity to the windowing configuration is analysed experimentally in Sec.~\ref{sec:Effect_windowing}.

The node feature vector is defined as
\begin{equation}
\mathbf{h}_i^{(t)}=
\Big[\,\log\!\big(\lvert \mathcal{F}\{x_i^{(t)}\}(f_k)\rvert+\varepsilon\big)\,\Big]_{k=1}^{d}
\in\mathbb{R}^{d},
\end{equation}
where \(\{f_k\}_{k=1}^{d}\subset\mathcal{B}\) are the retained one-sided frequency bins. The feature dimension \(d\), corresponding to the number of selected bins in the target frequency band, is determined by the frequency resolution induced by the window length \(S\) and the repetition time. This compact spectral representation captures slow BOLD oscillatory characteristics while avoiding an unnecessarily high-dimensional feature space in short windows. Stacking the node features over all ROIs yields the window-specific node-feature matrix \(X_t\in\mathbb{R}^{N\times d}\). 

\subsubsection{Functional Connectivity}
For each window \(t\), we compute the ROI-wise Pearson correlation coefficients from the windowed BOLD signals and apply the Fisher-\(z\) transform, obtaining a symmetric matrix \(\mathbf{C}^{(t)}\in\mathbb{R}^{N\times N}\) with entries
\begin{equation}
c_{ij}^{(t)}=\frac{1}{2}\log\frac{1+\rho_{ij}^{(t)}}{1-\rho_{ij}^{(t)}}.
\end{equation}
Because short-window Fisher-\(z\) values can vary substantially in dynamic range across windows, we apply window-wise min-max normalisation to the off-diagonal entries,
\begin{equation}
\widetilde{c}_{ij}^{(t)}
=
\frac{c_{ij}^{(t)}-c_{\min}^{(t)}}
{c_{\max}^{(t)}-c_{\min}^{(t)}+\epsilon},
\qquad i\neq j,
\end{equation}
where \(c_{\min}^{(t)}\) and \(c_{\max}^{(t)}\) are computed over the off-diagonal entries of \(\mathbf{C}^{(t)}\). This normalisation serves two purposes. First, it standardises the edge-weight scale across windows so that subsequent sparsification has a consistent interpretation. Second, it maps the graph operator to a nonnegative range, which is desirable for the weighted message-passing scheme used later. After this transformation, the resulting values should be interpreted as relative edge strengths within each window rather than statistically calibrated Fisher-\(z\) scores.

We then perform weighted sparsification:
\begin{equation}
\label{eq:tau}
\widehat{A}_{ij}^{(t)}
=
\widetilde{c}_{ij}^{(t)}\,
\mathbb{I}\big(\widetilde{c}_{ij}^{(t)}\ge\tau\big),
\qquad
\widehat{A}_{ii}^{(t)}=0,
\end{equation}
where \(\tau\) is a fixed threshold and \(\mathbb{I}(\cdot)\) denotes the indicator function, which equals \(1\) when its argument is true and \(0\) otherwise. This step removes weak and unstable short-window connections while preserving the magnitudes of retained edges. Importantly, thresholding is applied to the window-wise normalised values rather than to the raw Fisher-\(z\) scores. Therefore, although \(\tau\) is numerically fixed, it corresponds to a window-dependent cutoff in the original connectivity space:
\begin{equation}
c_{ij}^{(t)}
\ge
c_{\min}^{(t)}+\tau\bigl(c_{\max}^{(t)}-c_{\min}^{(t)}+\epsilon\bigr).
\end{equation}
Hence, \(\tau\) does not impose a globally fixed connectivity threshold across all windows and subjects. Instead, it retains edges that are relatively strong within each window, and thus primarily controls the relative sparsity of each window-specific graph. Since the same transformation is applied symmetrically to each ROI pair, \(\widehat{A}^{(t)}\) remains symmetric. The sensitivity to \(\tau\) is analysed experimentally in Sec.~\ref{sec:connectivity_threshold}.

\subsubsection{Distance-decay Prior}

The window-specific \(\widehat{A}^{(t)}\) is estimated from statistical similarity and does not explicitly encode spatial proximity in anatomical space. However, short-window correlations between spatially distant regions tend to be more variable and are more susceptible to residual motion and nuisance effects \cite{power2012spurious,korponay2024brain}. 
To introduce a spatial prior while preserving potentially meaningful long-range interactions, we construct a distance-decay prior based on ROI centroid distances, which serve as a simple and computationally efficient proxy for anatomical proximity.

We define the distance prior using the exponential kernel
\begin{equation}
D_{ij}=\exp(-d_{ij}/\sigma),
\end{equation}
where \(d_{ij}=\lVert \mathbf{r}_i-\mathbf{r}_j\rVert_2\) is the Euclidean distance between ROIs \(i\) and \(j\), \(\sigma=\operatorname{median}\{d_{ij}:1\le i<j\le N\}\) is a data-derived distance scale, and \(\mathbf{r}_i\in\mathbb{R}^{3}\) denotes the centroid of ROI \(i\) in MNI space \cite{evans2012brain}.
We adopt an exponential kernel for two reasons. First, compared with a Gaussian decay, it imposes a stronger penalty on moderately distant pairs, which is desirable when short-window DFC is particularly sensitive to unstable medium-to-long-range correlations. Second, it retains a heavier tail than a Gaussian kernel, so distant pairs are attenuated rather than abruptly suppressed. This is important because diagnostically relevant long-range functional interactions may still exist and should remain available to the model when supported by sufficiently strong empirical connectivity. Since \(D_{ij}>0\) for all ROI pairs, the distance prior modulates long-range edges but never hard-removes them. The ablation study in Sec.~\ref{sec:ablation_study} further indicates that incorporating this prior improves performance, supporting its role as a stabilising inductive bias. The reliability-refined adjacency is obtained by
\begin{equation}
A^{(t)}=\widehat{A}^{(t)}\odot D.
\end{equation}

The resulting graph therefore favours edges that are functionally strong while modulating them according to spatial proximity, thereby reducing the influence of noisy long-range correlations without hard-suppressing potentially meaningful large-scale interactions. By this construction, each subject is represented by a sequence of dynamic brain graphs \(\{\mathcal{G}_t(V,E_t)\}_{t=1}^{T}\), together with node-feature matrices \(X=\{X_t\}_{t=1}^{T}\) and adjacency matrices \(A=\{A^{(t)}\}_{t=1}^{T}\), where \(X_t\in\mathbb{R}^{N\times d}\) and \(A^{(t)}\in\mathbb{R}^{N\times N}\).

\subsection{Joint Spatio-Temporal Graph Learning}

To obtain stable window-wise ROI embeddings before temporal coupling, we first apply a local spatial encoder to each graph snapshot \(\mathcal{G}_t\). Specifically, a weighted GIN-style spatial encoder is used as a feature-lifting module that transforms the window-specific node-feature matrix \(X_t\) and reliability-refined adjacency \(A^{(t)}\) into latent ROI representations \(H_t\in\mathbb{R}^{N\times C}\). For layer \(l\), the update is
\begin{equation}
\!\!\! \mathbf{h}_{v}^{(l+1)}
=
\phi^{(l+1)} (
(1+\epsilon^{(l+1)})\,\mathbf{h}_{v}^{(l)}
+
\sum_{u\in\mathcal{N}(v)} A_{vu}^{(t)}\,\mathbf{h}_{u}^{(l)}
),
\label{eq:gin}
\end{equation}
where \(\phi^{(l+1)}\) is an MLP, \(\epsilon^{(l+1)}\) is learnable, and \(A_{vu}^{(t)}\) denotes the reliability-refined edge weight between ROIs \(u\) and \(v\) in window \(t\). In this way, the local spatial encoder incorporates both the graph topology and the window-specific edge strengths during feature lifting. Stacking the resulting window-wise features gives $H=\big[H_1,\dots,H_T\big]\in\mathbb{R}^{T\times N\times C}.$

The principal spatio-temporal representation learning step is then carried out on a joint graph over all time-ROI pairs \((t,i)\). The tensor \(H\) is reshaped as the initial node-feature matrix of the joint graph, where each node corresponds to a time-ROI pair \((t,i)\). The within-window graphs are placed on the diagonal blocks to form the spatial operator
\begin{equation}
\mathcal{A}_{\mathrm{sp}}
=
\operatorname{blkdiag}\!\big(A^{(1)},A^{(2)},\ldots,A^{(T)}\big)
\in\mathbb{R}^{(TN)\times(TN)}.
\end{equation}

To move beyond independent within-window connectivity modeling, we further construct a joint spatio-temporal graph by augmenting within-window functional connectivity with cross-window identity-preserving temporal links under an explicit Hawkes-inspired temporal dependency prior. Specifically, each ROI is connected to itself in future windows up to lag \(L\), and the strength of each temporal edge decays exponentially with the window lag. 
This design enables the model to capture directed temporal dependencies while preserving the within-window spatial connectivity structure, so that spatial and temporal dependencies can be propagated within a unified graph operator.
Formally, we instantiate this temporal prior by connecting the same ROI forward in time up to lag \(L\).
Let
\begin{equation}
\label{eq:delta}
w_{\Delta t}=\alpha e^{-\beta \Delta t},
\qquad \Delta t=1,\ldots,L,
\end{equation}
where \(\alpha,\beta>0\) are learnable scalar parameters and \(\Delta t\) denotes the window lag. Let \(\mathbf{J}_{\Delta t}\in\mathbb{R}^{T\times T}\) be the lag-\(\Delta t\) forward-shift matrix whose nonzero entries satisfy \([\mathbf{J}_{\Delta t}]_{t,t+\Delta t}=1\), and \(\mathbf{I}_N\) is the \(N\times N\) identity matrix. The temporal operator is
\begin{equation}
\mathcal{A}_{\mathrm{tmp}}
=
\sum_{\Delta t=1}^{L} w_{\Delta t}\,(\mathbf{J}_{\Delta t}\otimes \mathbf{I}_N) \in\mathbb{R}^{(TN)\times(TN)},
\end{equation}
where $\otimes$ denotes the Kronecker product, 
and the joint spatio-temporal graph operator is
$\bar{\mathbf{A}}
=
\mathcal{A}_{\mathrm{sp}}+\mathcal{A}_{\mathrm{tmp}}.$
Because the temporal shift matrices \(\mathbf{J}_{\Delta t}\) encode forward temporal links, the temporal operator is generally directed even though each within-window spatial block remains symmetric.

We then perform weighted message passing on this joint graph. The weighted STGIN operator is defined as
\begin{equation} \!\!\! 
\mathbf{h}_{v}^{(l+1)}
=
\phi^{(l+1)} (
(1+\epsilon^{(l+1)})\,\mathbf{h}_{v}^{(l)}
+
\sum_{u\in\mathcal{N}(v)} \bar{\mathbf{A}}_{vu}\,\mathbf{h}_{u}^{(l)}),
\label{eq:stgin}
\end{equation}
where \(\bar{\mathbf{A}}_{vu}\) explicitly weights both within-window spatial messages and cross-window temporal messages. Stacking several such layers produces the spatio-temporal node representations $Z\in\mathbb{R}^{T\times N\times C}.$
In this design, the initial window-wise encoder serves as a local feature-lifting module, whereas the weighted STGIN operator performs the principal spatio-temporal propagation under the proposed Hawkes-inspired temporal dependency prior.

\subsection{Self-supervised Hawkes-inspired Kernel-weighted Contrastive Learning (HW-CL)}
\label{HWCL}
The joint spatio-temporal graph encoder described above captures coupled within-window spatial structure and cross-window temporal dependencies. To further encourage discriminative and temporally coherent ROI representations before introducing label supervision, we impose a self-supervised Hawkes-inspired kernel-weighted contrastive objective, referred to as HW-CL. The key idea is to align the contrastive supervision with the same lag-dependent temporal prior used in the graph construction stage. In particular, temporally closer cross-window pairs are emphasized more strongly, while more distant pairs are assigned smaller weights according to the learnable exponential decay in Eq.~\eqref{eq:delta}. In this way, the pretraining objective is structurally consistent with the temporal dependency model embedded in the joint graph.

\begin{figure}[t]
\centering
\includegraphics[width=\columnwidth]{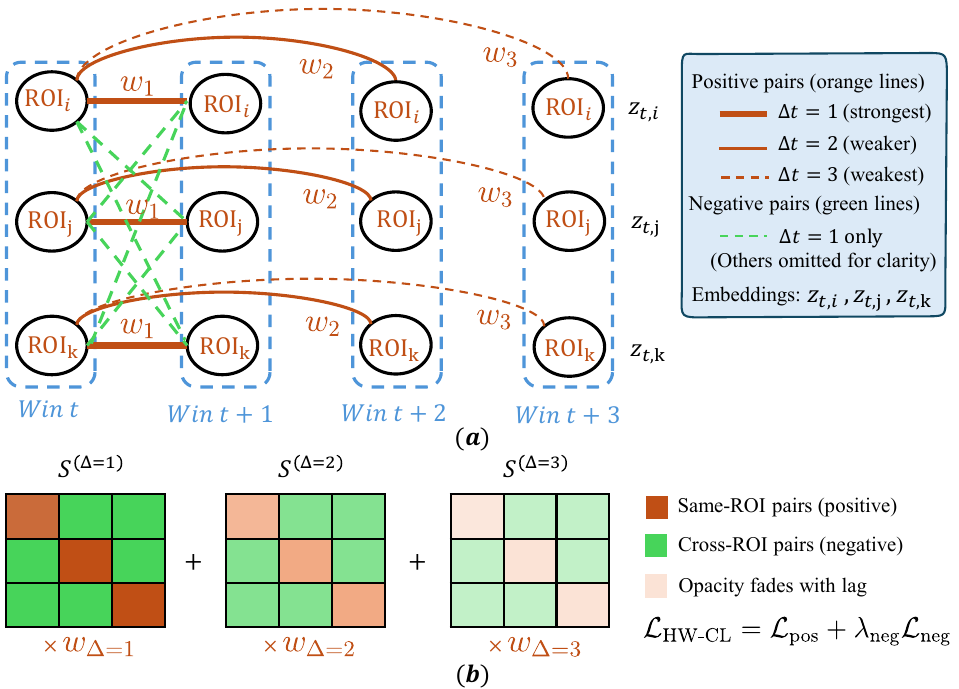}
\caption{Illustration of the Hawkes-inspired kernel-weighted contrastive objective.
(a) Contrastive pairing strategy: for each temporal lag \(\Delta\), same-ROI pairs across windows form positive pairs weighted by the Hawkes-inspired kernel \(w_\Delta\), with larger weights assigned to smaller lags. Different-ROI pairs form negative pairs, shown here only for \(\Delta=1\) to avoid visual clutter.
(b) Lag-wise cosine similarity matrices \(S^{(\Delta)}\): diagonal entries (positive pairs) are encouraged toward \(1\), whereas off-diagonal entries (negative pairs) are encouraged toward \(0\). The lag-wise errors are aggregated with Hawkes-inspired weighting to form \(\mathcal{L}_{\mathrm{HW\text{-}CL}}=\mathcal{L}_{\mathrm{pos}}+\lambda_{\mathrm{neg}}\mathcal{L}_{\mathrm{neg}}\), so that temporally closer windows contribute more strongly.}
\label{fig_hwcl}
\end{figure}

Given the spatio-temporal node representations \(Z=\{Z^{(t)}\in\mathbb{R}^{N\times C}\}_{t=1}^{T}\), we first apply \(\ell_2\) normalisation along the feature dimension,
$\widetilde{Z}_{t,n}
=
\frac{Z_{t,n}}{\lVert Z_{t,n}\rVert_2},$
where $t=1,\ldots,T,\ \ n=1,\ldots,N.$
For each lag \(\Delta\in\{1,\ldots,\min(L,T-1)\}\), we compute the lag-wise cosine similarities
\begin{equation}
S_{n,m}^{(\Delta)}(t)
=
\langle \widetilde{Z}_{t,n},\widetilde{Z}_{t+\Delta,m}\rangle,
\qquad
t=1,\ldots,T-\Delta.
\end{equation}
Diagonal entries \((n=m)\) correspond to positive pairs, namely temporally separated observations of the same ROI, whereas off-diagonal entries \((n\neq m)\) correspond to negative pairs formed by different ROIs across windows. As illustrated in Fig.~\ref{fig_hwcl}(a), each lag defines a cross-window pairing pattern, and the contribution of that lag is weighted by the same Hawkes-inspired decay term \(w_{\Delta}\).
Based on this construction, the positive-pair term is defined as
\begin{equation}
\mathcal{L}_{\mathrm{pos}}
=
\frac{
\displaystyle
\sum_{\Delta=1}^{\min(L,T-1)}
w_\Delta
\sum_{t=1}^{T-\Delta}\sum_{n=1}^{N}
\bigl(1-S_{n,n}^{(\Delta)}(t)\bigr)^2
}{
\displaystyle
\sum_{\Delta=1}^{\min(L,T-1)}
w_\Delta (T-\Delta)\,N
},
\end{equation}
which encourages temporally separated observations of the same ROI to remain similar in the learned latent space. The negative-pair term is defined as
\begin{equation}
\mathcal{L}_{\mathrm{neg}}
=
\frac{
\displaystyle
\sum_{\Delta=1}^{\min(L,T-1)}
w_\Delta
\sum_{t=1}^{T-\Delta}\sum_{n=1}^{N}
\sum_{\substack{m=1\\m\neq n}}^{N}
\bigl(S_{n,m}^{(\Delta)}(t)\bigr)^2
}{
\displaystyle
\sum_{\Delta=1}^{\min(L,T-1)}
w_\Delta (T-\Delta)\,N\,(N-1)
},
\end{equation} 
which suppresses redundant cross-window similarity between different ROIs.

The overall HW-CL objective is
\begin{equation}
\mathcal{L}_{\mathrm{HW\text{-}CL}}
=
\mathcal{L}_{\mathrm{pos}}
+
\lambda_{\mathrm{neg}}\mathcal{L}_{\mathrm{neg}},
\end{equation}
where \(\lambda_{\mathrm{neg}}>0\) balances the positive- and negative-pair terms. As shown in Fig.~\ref{fig_hwcl}(b), each lag induces a cosine similarity matrix whose diagonal and off-diagonal entries contribute to the positive- and negative-pair terms, respectively. The lag-wise losses are then aggregated using the Hawkes-inspired kernel weights so that temporally closer windows contribute more strongly to pretraining.

Because HW-CL directly penalizes pairwise similarity deviations through squared-error objectives on positive and negative pairs, no additional temperature parameter is required.
The lag weights appear in both the numerators and denominators, so that each term is computed as a lag-weighted average error. This preserves the relative emphasis on temporally closer pairs while avoiding changes in loss scale caused solely by the overall magnitude of the lag weights.

\subsection{Supervised Downstream Classification}
For subject-level diagnosis, we attach a lightweight downstream readout head to the pretrained spatio-temporal encoder. The purpose of this head is to adapt the compact and meaningful spatio-temporal embeddings learned during Stage~1 to the supervised diagnostic objective. Specifically, for each window \(t\), a ChebNet takes \(Z_t\) as node features and \(A^{(t)}\) as the graph operator to refine within-window graph representations. A GRU then aggregates the ordered sequence of window-level representations into a subject-level embedding, which is fed to a fully connected layer to predict the probability of MDD.
Let \(y_i\in\{0,1\}\) denote the ground-truth label for subject \(i\), and let \(\hat p_i\in(0,1)\) denote the predicted probability of the positive class. The classification loss is the binary cross-entropy
\begin{equation}
\mathcal{L}_{\mathrm{cls}}
=
-\frac{1}{B}\sum_{i=1}^{B}
\Big[
y_i\log\hat p_i
+
(1-y_i)\log(1-\hat p_i)
\Big].
\end{equation}

\subsection{Training Strategy}
The proposed framework is trained in two stages. In Stage~1, the spatial feature-lifting module, the weighted STGIN encoder, and the learnable temporal kernel parameters are pretrained using the HW-CL objective. The goal of this stage is to learn compact and meaningful ROI-level spatio-temporal embeddings before introducing label supervision. The importance of this pretraining stage is further examined in the ablation study. As shown by the w/o HW-CL setting in Table~\ref{tab:ablation_study} (Sec.~\ref{sec:ablation_study}), removing HW-CL results in inferior classification.

In Stage~2, the diagnostic readout head is attached, and the pretrained encoder is further refined rather than frozen. All encoder parameters, together with the downstream readout head, are optimized jointly under the downstream objective
\begin{equation}
\label{eq:total}
\mathcal{L}_{\mathrm{total}}
=
\mathcal{L}_{\mathrm{cls}}
+
\lambda_{\mathrm{HW}}\mathcal{L}_{\mathrm{HW\text{-}CL}},
\end{equation}
where $\lambda_{\mathrm{HW}}$ is the objective balancing coefficient that balances the HW-CL objective and the supervised classification objective. In this way, Stage~2 adapts the pretrained spatio-temporal embeddings to the classification task while retaining the complementary supervision provided by HW-CL. The sensitivity to $\lambda_{\mathrm{HW}}$ is analysed experimentally in Sec.~\ref{sec:Effect_lambda}.

\section{Experiments and Results}
\label{sec:exp_result}

\subsection{Experimental Setup}
\subsubsection{Datasets}
We used the publicly available REST-meta-MDD dataset, which includes participants with MDD and normal controls (NCs). To reduce preprocessing heterogeneity across sites, the REST-meta-MDD consortium processed the data using a standardized pipeline adapted from DPARSF~\cite{yan2010dparsf,Chen2022}.
Specifically, the first ten volumes were discarded, and the remaining images underwent motion correction, spatial normalization, spatial smoothing, and band-pass filtering (0.01--0.1~Hz). Following prior work~\cite{kong2024multi}, scans with fewer than 230 time points were excluded to reduce variability in scan length for dynamic functional connectivity analysis. The final sample comprised 368 subjects with MDD and 299 NCs. Whole-brain parcellation was performed using the AAL atlas, resulting in \(N=116\) ROIs~\cite{Rolls2020}.

\subsubsection{Implementation Details}
We evaluated the proposed method using subject-level 10-fold cross-validation over 10 random seeds, and report the mean and standard deviation across runs. Performance was assessed using accuracy (ACC), sensitivity (SEN), specificity (SPEC), F1-score (F1), and area under the receiver operating characteristic curve (AUC). All experiments were conducted on an NVIDIA RTX~4090 GPU using PyTorch~2.5.1. Dynamic graphs were constructed with \(T=8\) non-overlapping windows and threshold \(\tau=0.44\). The feature extractor consisted of a 2-layer SpatialGIN with hidden size 64, followed by a 3-layer STGIN with maximum temporal lag 3. In Stage~1, pretraining used AdamW with batch size 16, learning rate \(9\times10^{-4}\), weight decay \(6\times10^{-6}\), 40 epochs, and \(\lambda_{\mathrm{neg}}=0.2\). In Stage~2, the pretrained encoder and downstream classifier were jointly optimized. The downstream module consisted of a ChebNet (\(K=2\), hidden size 64, embedding dimension 256, dropout 0.2) and a bidirectional GRU with 256 hidden units. Joint optimization used AdamW with batch size 16, learning rate \(2\times10^{-3}\), weight decay \(4\times10^{-5}\), and 120 epochs. We set \(\lambda_{\mathrm{HW}}=0.1\) in all experiments. Sensitivity analyses for \(\tau\), \(T\), and \(\lambda_{\mathrm{HW}}\) are given in Sec.~\ref{sec:parameter_sensitivity}.

\subsection{Evaluation of Diagnostic Performance for MDD}
We compare HWSTCL with thirteen baselines, grouped into SFC and DFC methods. The SFC baselines include two conventional classifiers, support vector machine (SVM) and random forest (RF), and five deep models: BrainGNN~\cite{li2021braingnn}, IBGNN~\cite{cui2022interpretable}, GAE-FCNN~\cite{Noman2024}, BrainNetCNN~\cite{Kawahara2017}, and GNNMA~\cite{si2025graph}. Each uses a single time-averaged functional connectome per subject, with the upper-triangular entries of the static connectivity matrix vectorised as input. The DFC baselines include STGCN~\cite{kong2021spatio}, STAGCN~\cite{wang2022spatio}, ESTA~\cite{lee2024eigendecomposition}, Temporal-BCGCN~\cite{zhu2024temporal}, MSSTAN~\cite{kong2024multi} and Long-Interval STGC~\cite{li2025long}. These methods model rs-fMRI as sequences of graphs or temporal segments to capture both spatial topology and temporal variation in connectivity. In the following tables, the best and second-best results are shown in bold and underlined, respectively.

Table~\ref{tab:comp_sota} reports the mean and standard deviation over repeated 10-fold cross-validation runs for ACC, SEN, SPEC, F1, and AUC. Overall, HWSTCL achieves the best performance on all five metrics, with 70.43\% ACC, 76.30\% SEN, 60.96\% SPEC, 73.20\% F1, and 67.63\% AUC. Compared with the strongest DFC baselines, HWSTCL improves ACC by 1.78 points over MSSTAN, SEN by 1.61 points over STGCN, SPEC by 1.51 points over MSSTAN, F1 by 1.34 points over Temporal-BCGCN, and AUC by 0.57 points over MSSTAN. These results suggest that the proposed framework improves classification performance while providing a more balanced trade-off between sensitivity and specificity. A clear performance hierarchy can be observed in Table~\ref{tab:comp_sota}. Conventional classifiers such as SVM and RF achieve the lowest scores, suggesting that shallow models are insufficient to capture the complex and nonlinear structure of rs-fMRI connectivity patterns. Static graph-based methods improve substantially over these baselines, confirming the value of connectome topology for MDD diagnosis. However, they still perform worse than the stronger DFC methods, suggesting that explicitly modeling temporal variation in connectivity is beneficial for this task.


Among the DFC baselines, methods such as STGCN, STAGCN, MSSTAN, and Long-Interval STGC already demonstrate the advantage of incorporating temporal information. Nevertheless, their gains remain limited relative to HWSTCL. A plausible explanation is that these methods do not jointly address the three issues emphasized in this work: the reliability of short-window connectivity estimation, the informativeness of node representations, and the alignment between temporal coupling and the Hawkes-inspired kernel-weighted contrastive objective. In contrast, HWSTCL combines distance-refined graph construction, frequency-informed ROI descriptors, joint spatio-temporal message passing under a Hawkes-inspired temporal dependency prior, and Hawkes-inspired kernel-weighted contrastive pretraining. This combination appears to produce more discriminative and more stable spatio-temporal representations, which are then further refined during downstream joint optimization for MDD classification.

\begin{table*}[htbp]
\centering
\caption{Performance comparison of competing methods and the proposed approach for MDD vs.\ NC on the benchmark MDD dataset using the AAL atlas. Results are reported as mean~$\pm$~standard deviation.}
\label{tab:comp_sota}
\renewcommand{\arraystretch}{1.15}
\setlength{\tabcolsep}{5pt}
\begin{tabularx}{\textwidth}{@{}ll*{5}{>{\centering\arraybackslash}X}@{}}
\toprule[1pt]
\multicolumn{2}{l}{Method} &
\multicolumn{1}{c}{ACC (\%)} &
\multicolumn{1}{c}{SEN (\%)} &
\multicolumn{1}{c}{SPEC (\%)} &
\multicolumn{1}{c}{F1 (\%)} &
\multicolumn{1}{c}{AUC (\%)} \\
\midrule
\multirow{7}{*}{\emph{SFC}} 
& SVM         & \mstd{56.60}{0.75} & \mstd{60.87}{1.08} & \mstd{51.51}{0.98} & \mstd{60.49}{0.85} & \mstd{56.19}{0.91} \\
& RF          & \mstd{54.67}{1.59} & \mstd{62.18}{1.23} & \mstd{45.91}{1.54} & \mstd{53.91}{1.58} & \mstd{53.91}{1.58} \\
& BrainGNN \cite{li2021braingnn}    & \mstd{65.18}{0.99} & \mstd{73.74}{1.80} & \mstd{52.91}{5.23} & \mstd{69.11}{1.80} & \mstd{63.32}{0.92} \\
& IBGNN \cite{cui2022interpretable} & \mstd{63.28}{1.48} & \mstd{73.79}{3.18} & \mstd{50.28}{4.54} & \mstd{69.11}{2.12} & \mstd{62.28}{1.45} \\
& GAE-FCNN \cite{Noman2024}         & \mstd{66.19}{4.69} & \mstd{74.60}{3.90} & \mstd{54.84}{4.16} & \mstd{70.30}{4.90} & \mstd{64.50}{1.40} \\
& BrainNetCNN \cite{Kawahara2017}   & \mstd{64.40}{1.15} & \mstd{72.90}{2.50} & \mstd{51.70}{1.90} & \mstd{68.50}{1.35} & \mstd{62.40}{1.30} \\
& GNNMA \cite{si2025graph}          & \mstd{65.70}{1.00} & \mstd{74.05}{2.10} & \mstd{54.20}{3.30} & \mstd{69.70}{1.20} & \mstd{63.80}{1.10} \\
\midrule
\multirow{6}{*}{\emph{DFC}}
& STGCN \cite{kong2021spatio}       & \mstd{61.21}{1.00} & \underline{\mstd{74.69}{4.25}} & \mstd{40.71}{5.03} & \mstd{65.35}{2.30} & \mstd{48.81}{2.14} \\
& STAGCN \cite{wang2022spatio}      & \mstd{66.04}{0.66} & \mstd{74.45}{1.92} & \mstd{53.37}{5.27} & \mstd{69.55}{1.82} & \mstd{58.80}{0.61} \\
& ESTA \cite{lee2024eigendecomposition} 
                                      & \mstd{65.92}{1.92} & \mstd{73.23}{2.56} & \mstd{57.10}{4.99} & \mstd{70.04}{1.46} & \mstd{63.62}{1.14} \\
& Temporal-BCGCN \cite{zhu2024temporal} 
                                      & \mstd{68.01}{0.92} & \mstd{74.20}{2.80} & \mstd{58.80}{1.10} & \underline{\mstd{71.86}{1.15}} & \mstd{64.72}{1.04} \\
& MSSTAN \cite{kong2024multi}                             
                                      & \underline{\mstd{68.65}{0.90}} 
                                      & \mstd{74.67}{3.26} 
                                      & \underline{\mstd{59.45}{4.84}} 
                                      & \mstd{71.63}{1.23} 
                                      & \underline{\mstd{67.06}{1.39}} \\
& Long-Interval STGC \cite{li2025long}
                                      & \mstd{67.53}{0.95} & \mstd{73.52}{2.65} & \mstd{57.91}{4.84} & \mstd{70.69}{1.30} & \mstd{65.09}{1.39} \\
\midrule
& HWSTCL (Ours)                      
                                      & \textbf{\mstd{70.43}{1.12}} 
                                      & \textbf{\mstd{76.30}{2.55}} 
                                      & \textbf{\mstd{60.96}{3.56}} 
                                      & \textbf{\mstd{73.20}{1.01}} 
                                      & \textbf{\mstd{67.63}{1.23}} \\
\bottomrule
\end{tabularx}
\end{table*}

\subsection{Ablation Study}
\label{sec:ablation_study}
We conducted an ablation study to evaluate the main components of HWSTCL, including the distance-decay prior (\(D\)), the local spatial encoder (GIN), joint spatio-temporal propagation (Joint ST), the HW-CL objective, and the node feature design.

\subsubsection{Ablation on Core Components}

Four variants were evaluated to examine the main components of HWSTCL. ``w/o \(D\)'' removes the distance-decay prior from the Pearson-correlation graphs. ``w/o GIN'' removes the local spatial encoder before joint spatio-temporal propagation. ``w/o Joint ST'' removes temporal propagation and retains only window-wise spatial encoding. ``w/o HW-CL'' removes HW-CL pretraining in Stage 1 while retaining the Stage 2 training objective.
Table~\ref{tab:ablation_study} summarizes the ablation results. The full HWSTCL model achieves the best overall performance, with the highest ACC, SEN, F1, and AUC, and the second-highest SPEC. These results indicate that the final performance benefits from the combined use of reliability-aware graph construction, local spatial feature lifting, joint spatio-temporal propagation, and the HW-CL objective.

\begin{table}[t]
\vspace{-0.15in}
\caption{Component ablation on the benchmark dataset.}
\label{tab:ablation_study}
\setlength{\tabcolsep}{3pt}
\renewcommand{\arraystretch}{1.15}
\resizebox{\columnwidth}{!}{%
\begin{tabular}{lccccc}
\hline
Component & ACC (\%) & SEN (\%) & SPEC (\%) & F1 (\%) & AUC (\%) \\
\hline
HWSTCL & \textbf{70.43\,$\pm$\,1.12} & \textbf{76.30\,$\pm$\,2.55} & \secbest{60.96\,$\pm$\,3.56} & \textbf{73.20\,$\pm$\,1.01} & \textbf{67.63\,$\pm$\,1.23} \\
w/o \(D\) & 68.60\,$\pm$\,1.10 & 74.00\,$\pm$\,1.20 & 60.50\,$\pm$\,1.15 & \secbest{72.10\,$\pm$\,1.10} & 66.45\,$\pm$\,1.12 \\
w/o GIN & 68.23\,$\pm$\,1.36 & 73.60\,$\pm$\,1.85 & 59.26\,$\pm$\,1.72 & 70.25\,$\pm$\,1.46 & 65.82\,$\pm$\,1.42 \\
w/o Joint ST & \secbest{68.65\,$\pm$\,1.10} & \secbest{74.80\,$\pm$\,1.15} & \textbf{61.00\,$\pm$\,1.20} & 71.40\,$\pm$\,1.08 & \secbest{67.55\,$\pm$\,1.14} \\
w/o HW-CL & 66.54\,$\pm$\,1.25 & 71.56\,$\pm$\,1.81 & 59.76\,$\pm$\,1.28 & 70.16\,$\pm$\,1.76 & 66.04\,$\pm$\,1.85 \\
\hline
\end{tabular}}
\end{table}

\begin{table}[t]
\vspace{-0.15in}
\caption{Influence of node features on HWSTCL.}
\label{Table_IV}
\setlength{\tabcolsep}{3pt}
\renewcommand{\arraystretch}{1.15}
\resizebox{\columnwidth}{!}{%
\begin{tabular}{lccccc}
\hline
Node Feature   & ACC (\%)                 & SEN (\%)                  & SPEC (\%)                  & F1 (\%)                   & AUC (\%)                  \\
\hline
FFT        & \textbf{70.43\,$\pm$\,1.12} & \textbf{76.30\,$\pm$\,2.55}  & \textbf{60.96\,$\pm$\,3.56}  & \secbest{73.20\,$\pm$\,1.01}  & \secbest{67.63\,$\pm$\,1.23} \\
PSD         & 67.12\,$\pm$\,1.09          & 70.10\,$\pm$\,1.20           & 59.80\,$\pm$\,1.15           & 71.90\,$\pm$\,1.08           & 66.80\,$\pm$\,1.10          \\
RBS       & 67.40\,$\pm$\,1.14          & 71.50\,$\pm$\,1.18           & 60.10\,$\pm$\,1.18           & 72.10\,$\pm$\,1.12           & \textbf{68.10\,$\pm$\,1.17}          \\
Correlation     & \secbest{68.90\,$\pm$\,1.11} & \secbest{73.20\,$\pm$\,1.16}  & \secbest{60.36\,$\pm$\,1.16}  & \textbf{73.85\,$\pm$\,1.09}  & 66.40\,$\pm$\,1.13 \\
Avg. BOLD        & 68.05\,$\pm$\,1.12          & 69.80\,$\pm$\,1.15           & 59.90\,$\pm$\,1.14           & 71.50\,$\pm$\,1.10           & 67.60\,$\pm$\,1.12          \\
\hline
\end{tabular}}
\end{table}

Removing \(D\) (w/o \(D\)) degrades all metrics, showing that distance-aware refinement improves the robustness of short-window graph construction by suppressing unreliable long-range associations. 
Removing the local spatial encoder (w/o GIN) also reduces performance, suggesting that stronger local spatial feature lifting is beneficial before temporal coupling.

The contribution of joint spatio-temporal propagation is also evident. Compared with ``w/o Joint ST'', the full model improves ACC, SEN, F1, and AUC, suggesting that propagation on the joint spatio-temporal graph provides additional information beyond independent window-wise spatial encoding. Although ``w/o Joint ST'' achieves the highest SPEC, its lower ACC, F1, and AUC indicate that the full model provides a more favourable overall trade-off across metrics.

Removing HW-CL pretraining also leads to a reduction in all reported metrics. This result suggests that the HW-CL objective provides a useful pretraining signal for representation learning. In particular, the decrease in ACC, SEN, F1, and AUC suggests that HW-CL helps the encoder learn temporally coherent ROI representations that remain beneficial during downstream joint optimization.

\subsubsection{Ablation on Node Feature Design}
We further compared FFT-based spectral features with raw BOLD signals (RBS), Welch power spectral density (PSD), correlation-derived node features, and per-window mean BOLD features. This analysis evaluates whether the proposed frequency-informed node representation contributes to the effectiveness of HWSTCL. 
As shown in Table~\ref{Table_IV}, FFT-based node features achieve the best ACC, SEN, and SPEC, while also providing competitive F1 and AUC performance. Although correlation-derived features attain the highest F1 and RBS achieves the highest AUC, FFT features provide the strongest overall trade-off across the reported metrics. This result supports the use of frequency-informed ROI descriptors in HWSTCL and suggests that spectral characterization of regional BOLD dynamics provides a useful complement to connectivity-based graph construction.

\subsection{Sensitivity Analysis}
\label{sec:parameter_sensitivity}

\begin{figure*}[t]
\centering

\mysubfig{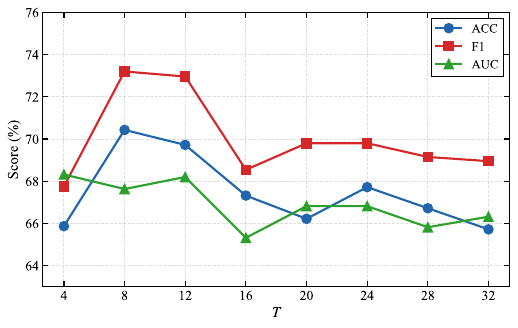}{Temporal window $T$ in Sec.~\ref{sec:node_f}.}{fig:sensitivity:a}
\hfill
\mysubfig{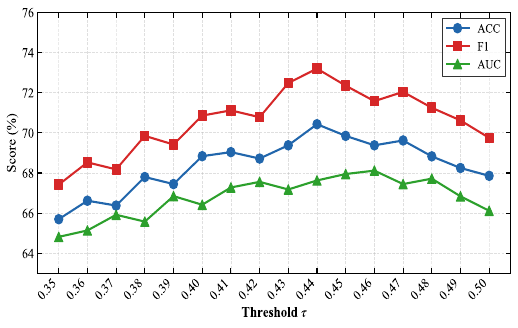}{Connectivity threshold $\tau$ in Eq.~\ref{eq:tau}.}{fig:sensitivity:b}
\hfill
\mysubfig{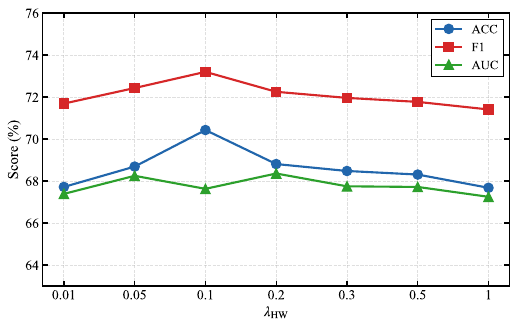}{Weight $\lambda_{HW}$ in Eq.~\ref{eq:total}.}{fig:sensitivity:c}

\caption{Parameter sensitivity analysis. The curves show the mean ACC, F1, and AUC over repeated 10-fold cross-validation runs.}
\label{fig:sensitivity}
\end{figure*}

We further examined the sensitivity of HWSTCL to several key hyperparameters involved in dynamic functional connectivity construction and model training, including the number of temporal windows \(T\), the sparsification threshold \(\tau\), and the objective balancing coefficient \(\lambda_{\mathrm{HW}}\). In each analysis, one parameter was varied while the others were fixed at their default settings. The results are summarized in terms of ACC, F1, and AUC in Fig.~\ref{fig:sensitivity}, where the mean performance and the corresponding standard deviation are reported.

\subsubsection{Effect of Temporal Window Length}
\label{sec:Effect_windowing}
We evaluated eight windowing settings with \(T\) ranging from 4 to 32 in order to study the impact of temporal granularity on dynamic connectivity modeling. As shown in Fig.~\ref{fig:sensitivity}(\ref{fig:sensitivity:a}), HWSTCL achieved the best overall performance at $T\!=\!8$. When \(T\) is too small, each window spans a relatively long temporal segment, which may smooth out transient connectivity variations and weaken the dynamic characterization of brain interactions. In contrast, excessively large \(T\) leads to shorter windows and less reliable connectivity estimation within each segment. These results suggest that $T\!=\!8$ provides a favorable balance between temporal sensitivity and connectivity stability.

\subsubsection{Effect of the Connectivity Threshold}
\label{sec:connectivity_threshold}

We next investigated the influence of the connectivity threshold \(\tau\), which controls graph sparsity during the thresholding of the normalized per-window correlation matrices. Specifically, \(\tau\) was varied from 0.35 to 0.50 with a step size of 0.01. As illustrated in Fig.~\ref{fig:sensitivity}(\ref{fig:sensitivity:b}), the best performance was obtained around $\tau\!=\!0.44$. A relatively small threshold retains too many weak connections and may introduce noisy or less informative interactions, whereas an overly large threshold removes useful connectivity patterns and results in excessively sparse graphs. The observed trend indicates that a moderate sparsity level is most suitable for preserving discriminative connectivity structure while maintaining effective message passing on the graph.

\subsubsection{Effect of the Objective Balancing Coefficient \(\lambda_{\mathrm{HW}}\)}
\label{sec:Effect_lambda}

Finally, we assessed the effect of the objective balancing coefficient \(\lambda_{\mathrm{HW}}\), which balances the Hawkes-inspired kernel-weighted contrastive objective and the supervised classification objective during training. We considered seven values ranging from 0.01 to 1.0. As shown in Fig.~\ref{fig:sensitivity}(\ref{fig:sensitivity:c}), the best performance was achieved at \(\lambda_{\mathrm{HW}}=0.1\). This result indicates that the Hawkes-inspired kernel-weighted contrastive objective is beneficial when properly balanced against the supervised classification objective. When \(\lambda_{\mathrm{HW}}\) becomes too large, the contrastive objective may dominate optimization and lead to degraded classification performance.

\subsection{Effect of the Downstream Classifier}
We used ChebNet\,+\,GRU as the default downstream classifier and compared it with the alternative classifiers in Table~\ref{Table_V}. As shown in Table~\ref{Table_V}, ChebNet\,+\,GRU achieved the best overall and most balanced performance, with the highest ACC, SEN, F1, and AUC, whereas ChebNet\,+\,TCN obtained a slightly higher SPEC. These results suggest that the proposed framework remains effective across different downstream readout designs, while ChebNet\,+\,GRU provides the strongest overall performance in our setting.
\begin{table}[h]
\caption{Influence of diagnostic classifiers on classification.}
\label{Table_V}
\setlength{\tabcolsep}{3pt}
\renewcommand{\arraystretch}{1.15}
\resizebox{\columnwidth}{!}{%
\begin{tabular}{lccccc}
\hline
Classifier   & ACC (\%) & SEN (\%) & SPEC (\%) & F1 (\%) & AUC (\%) \\
\hline
MLP            & 64.50\,$\pm$\,1.50 & 68.20\,$\pm$\,2.80 & 55.10\,$\pm$\,3.40 & 66.30\,$\pm$\,1.60 & 61.90\,$\pm$\,1.70 \\
ChebNet\,+\,MLP  & 66.20\,$\pm$\,1.30 & 71.00\,$\pm$\,2.50 & 58.00\,$\pm$\,3.20 & 69.10\,$\pm$\,1.40 & 64.80\,$\pm$\,1.50 \\
ChebNet\,+\,LSTM & \secbest{69.10\,$\pm$\,1.08} & \secbest{75.40\,$\pm$\,2.20} & 60.80\,$\pm$\,3.10 & 72.85\,$\pm$\,0.95 & \secbest{67.30\,$\pm$\,1.15} \\
ChebNet\,+\,TCN  & 69.00\,$\pm$\,1.05 & 75.10\,$\pm$\,2.10 & \textbf{61.25\,$\pm$\,3.00} & \secbest{72.95\,$\pm$\,0.98} & 66.60\,$\pm$\,1.10 \\
ChebNet\,+\,RNN  & 68.80\,$\pm$\,1.15 & 74.90\,$\pm$\,2.35 & 60.20\,$\pm$\,3.25 & 72.40\,$\pm$\,1.05 & 65.95\,$\pm$\,1.20 \\
ChebNet\,+\,GRU  & \textbf{70.43\,$\pm$\,1.12} & \textbf{76.30\,$\pm$\,2.55} & \secbest{60.96\,$\pm$\,3.56} & \textbf{73.20\,$\pm$\,1.01} & \textbf{67.63\,$\pm$\,1.23} \\
\hline
\end{tabular}}
\end{table}

\subsection{Interpretability of Discriminative Connectivity \& ROIs}
\label{sec:brain_conn}
\begin{figure}[t]
\centerline{\includegraphics[width=\columnwidth]{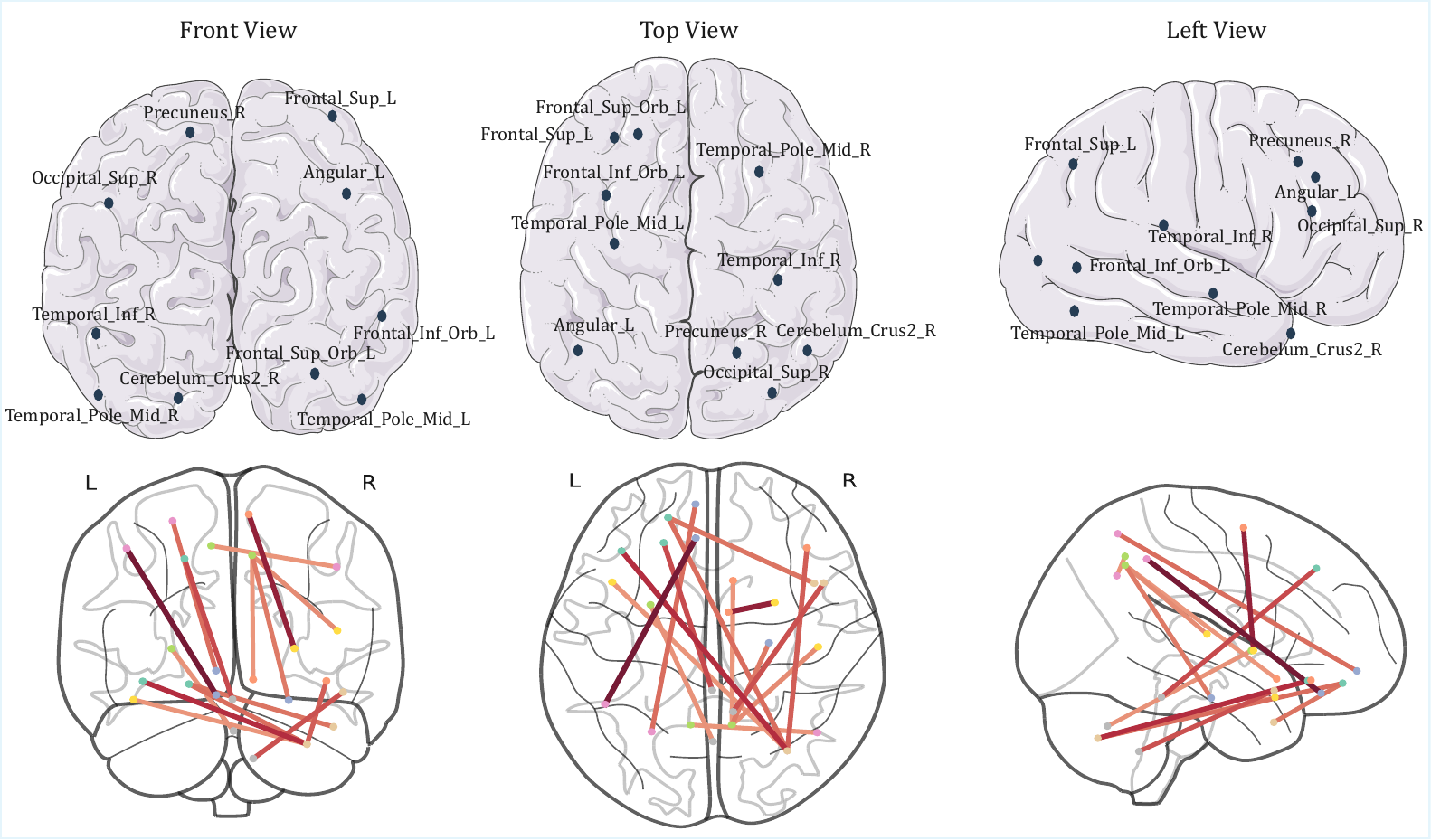}}
\caption{Top 10 brain regions (upper row) and top 15 most discriminative connections (lower row) identified by HWSTCL.}
\label{fig2}
\end{figure}

We further examined whether HWSTCL identifies neurobiologically meaningful connectivity patterns and brain regions for MDD diagnosis. To this end, we analyzed gradient-based saliency scores to identify the most important connections and regions for MDD diagnosis. Specifically, we report the top 15 most salient connections and summarize region-level importance by aggregating the saliency scores of incident connections for each ROI, thereby identifying the top 10 most salient regions, as shown in Fig.~\ref{fig2}. The most salient connections involve prefrontal regions, including ventromedial and orbitofrontal areas, together with cerebellar regions such as Crus~II. Additional high-importance links are also observed among default mode network hubs, including the precuneus, angular cortex, and parahippocampal regions. These findings are consistent with prior evidence of MDD-related abnormalities in default mode network connectivity~\cite{zhukovsky2021coordinate}, 
prefrontal circuits associated with affective and reward processing~\cite{hiser2018multifaceted}, 
cerebellar interactions with large-scale cortical systems~\cite{wang2023disrupted}, and broader MDD-related brain abnormalities~\cite{schmaal2016subcortical}.

Importantly, although the proposed distance-aware prior down-weights spatially distant connections, the highlighted patterns still include anatomically distributed large-scale interactions spanning prefrontal, cerebellar, and default mode network regions. This observation supports the intended role of the prior: rather than suppressing meaningful long-range functional coupling, it attenuates unreliable long-range correlations while preserving diagnostically relevant large-scale interactions when supported by sufficiently strong empirical connectivity. From the view of the overall framework, these results suggest that HWSTCL learns not only more discriminative representations for classification, but also more neurobiologically plausible spatio-temporal connectivity patterns.

\section{Discussion}

In this study, we proposed HWSTCL, a reliability-refined joint spatio-temporal graph contrastive learning framework for MDD diagnosis from rs-fMRI. The experimental results indicate that DFC-based diagnosis benefits from jointly improving graph construction and spatio-temporal representation learning. Compared with static FC methods, HWSTCL explicitly models temporal variations in functional connectivity, which may capture disease-related dynamic patterns that are averaged out in static connectomes. Compared with existing DFC baselines, the results suggest that temporal modeling alone may be insufficient when the underlying short-window graphs remain noisy or weakly regularised. The distance-decay prior provides a simple but effective anatomical constraint that reduces the influence of potentially unstable long-range correlations while preserving long-range interactions supported by strong empirical connectivity. The Hawkes-inspired temporal kernel further contributes by introducing an explicit lag-dependent dependency prior into both graph construction and contrastive learning. Rather than treating spatial graph learning and temporal aggregation as separate procedures, HWSTCL connects the same ROI across nearby future windows and performs message passing over a joint spatio-temporal graph. The HW-CL objective aligns the self-supervised learning signal with the same temporal dependency prior, encouraging ROI-anchored representations to remain consistent across nearby windows while suppressing redundant similarity between different ROIs. The ablation results support the importance of this design, as removing HW-CL or joint spatio-temporal propagation leads to degraded diagnostic performance. The interpretability analysis further suggests that HWSTCL captures neurobiologically meaningful patterns, with the most salient regions and connections involving prefrontal, orbitofrontal, default mode network, parahippocampal, and cerebellar areas, consistent with prior findings on MDD-related abnormalities in affective regulation and large-scale brain network organisation.

From a clinical perspective, HWSTCL should be regarded as an imaging-assisted decision-support framework rather than a standalone diagnostic tool, given the inherent complexity of MDD diagnosis arising from symptom heterogeneity, comorbidity, and medication effects. The value of the proposed framework lies in its ability to learn reproducible spatio-temporal connectivity representations from rs-fMRI that may complement symptom-based assessment. Beyond MDD, the proposed approach may also be applicable to other neurological and psychiatric disorders associated with altered functional connectivity dynamics, such as attention deficit hyperactivity disorder, autism spectrum disorder, mild cognitive impairment, and Parkinson's disease. Since many studies of these disorders also rely on short-window connectivity estimation, the reliability-aware graph construction and temporally coupled representation learning strategy introduced in HWSTCL may provide transferable methodological benefits. Looking ahead, several directions may further extend this work, including cross-site validation with domain adaptation strategies to evaluate generalisability across independent datasets~\cite{wang2026adhd}, incorporation of complementary structural information from diffusion tensor imaging or T1-weighted MRI for multi-modal graph construction, and privacy-preserving training paradigms such as federated learning~\cite{jiao2025fglfa} to support collaborative model development across institutions.

\section{Conclusion}
This study proposes HWSTCL, a Hawkes-inspired kernel-weighted spatio-temporal representation learning framework for MDD diagnosis from resting-state fMRI. It addresses three key challenges in DFC analysis: unreliable short-window connectivity estimation, insufficiently informative node representation, and decoupled spatio-temporal modeling. Accordingly, HWSTCL integrates reliability-aware graph construction, frequency-informed ROI descriptors, joint spatio-temporal propagation under a Hawkes-inspired temporal dependency prior, and a Hawkes-inspired kernel-weighted contrastive objective for ROI-anchored temporal consistency. Experimental results on the benchmark dataset show that HWSTCL outperforms representative static and dynamic baselines, supporting the effectiveness of reliability-refined joint spatio-temporal graph learning for rs-fMRI-based MDD diagnosis.

\bibliographystyle{IEEEtran}
\bibliography{references}
\end{document}